# The Qiyas Benchmark: Measuring ChatGPT Mathematical and Language Understanding in Arabic

Shahad Al-Khalifa
iWAN Research Group, College of Computer and Information Sciences
King Saud University
Riyadh, Saudi Arabia
shahadalkhalifa90@gmail.com

Hend Al-Khalifa
Department of Information Technology, College of Computer and Information Sciences
King Saud University
Riyadh, Saudi Arabia
hendk@ksu.edu.sa

*Abstract*— Despite the growing importance of Arabic as a global language, there is a notable lack of language models pre-trained exclusively on Arabic data. This shortage has led to limited benchmarks available for assessing language model performance in Arabic. To address this gap, we introduce two novel benchmarks designed to evaluate models' mathematical reasoning and language understanding abilities in Arabic. These benchmarks are derived from a General Aptitude Test (GAT) called Qiyas exam, a standardized test widely used for university admissions in Saudi Arabia. For validation purposes, we assess the performance of ChatGPT-3.5-trubo and ChatGPT-4 on our benchmarks. Our findings reveal that these benchmarks pose a significant challenge, with ChatGPT-4 achieving an overall average accuracy of 64%, while ChatGPT-3.5-trubo achieved an overall accuracy of 49% across the various question types in the Qiyas benchmark. We believe the release of these benchmarks will pave the way for enhancing the mathematical reasoning and language understanding capabilities of future models tailored for the low-resource Arabic language.

## I. INTRODUCTION

Evaluating the capabilities of large language models (LLMs) across different tasks like mathematical reasoning and natural language understanding is a critical area of research as these general-purpose AI systems become more widely used. Developing comprehensive evaluation benchmarks, especially for languages beyond English, is crucial for driving model improvement.

Arabic, a Semitic language with complex morphology and written from right-to-left, is spoken by over 400 million people across the Arab world [1]. Despite its status as a global language of importance, Arabic is considered low-resource in the field of natural language processing [2]. There is currently a shortage of LLMs pre-trained exclusively on large Arabic datasets. This has resulted in limited benchmarks available for robustly assessing Arabic LLM performance. While some prior work has translated English benchmarks, there is a need for high-quality, natively developed Arabic evaluation resources.

The lack of robust, Arabic-native benchmarks focused on key capabilities like math reasoning and language understanding represents a significant gap in the field. Having professionally designed evaluation resources in this domain could accelerate the development of higher-performing Arabic language models. This work makes two key contributions to address this need:

*1)* It introduces Qiyas, a benchmark suite for comprehensively evaluating LLM performance on mathematical and language tasks in the Arabic language. Qiyas consists of two components - a quantitative section assessing math skills and a verbal section evaluating Arabic language understanding abilities.

*2)* Using the Qiyas benchmark, the performance of the latest ChatGPT models (versions 3.5-turbo and 4) is extensively evaluated across zero-shot, one-shot, and few-shot settings to establish strong Arabic language baselines.

These benchmarks are derived from the Qiyas exam, a standardized exam widely used for university admissions in Saudi Arabia, ensuring their quality has been validated by educational experts. The results shed light on current LLM limitations for the Arabic language and highlight the impact of varied training data and prompting approaches. The release of Qiyas paves the way for advancing Arabic LLMs' mathematical reasoning and language understanding capabilities on these challenging, nationally representative Arabic tasks.

The remainder of the paper is structured as follows: The 'Background and Literature Review' section provides an overview of the Qiyas exam and discusses previous efforts in evaluating large language models (LLMs) in standardized exams. The 'Dataset Description' section describes the dataset used in the study. The 'Evaluation Approach' section outlines the methodology used to evaluate the performance of ChatGPT on the quantitative section assessing math skills and the verbal section evaluating Arabic language understanding abilities. The 'Results and Discussion' section delves into the benchmark results, analyzing the performance of both ChatGPT-3.5-turbo and ChatGPT-4 models. Finally, the 'Conclusion' section summarizes the key findings of the paper and provides an outlook on future research.

## II. BACKGROUND AND LITERATURE REVIEW

The National Centre for Assessment (aka Qiyas) is a significant institution in education and assessment. It is responsible for conducting standardized tests to assess the scholastic achievement of students applying for universities [3].

Qiyas is responsible for developing and implementing over 90 standardized and professional tests for the public and private sectors. It has over 1,500 test models and an item bank of over 230,000 questions. The tests consist of two sections: the verbal and the quantitative, focusing on students' analytical and deductive skills, helping them assess their learning capacity. The center also provides linguistic tests, including the English language efficiency test and the Arabic language test for non-native speakers. Additionally, it presents an assessment test for talented and creative students, as well as vocational tests, the most important of which is the Vocational Standards Test for Teachers [4].

Recent studies have evaluated the performance of large language models like ChatGPT on standardized exams across different domains. In the medical domain, ChatGPT has shown promising results, with studies indicating that it has reached the standard of passing third-year medical student exams [5]. Furthermore, research has demonstrated ChatGPT's success in passing the gold-standard US medical exam, suggesting significant potential applications in medicine [6]. Additionally, ChatGPT has been compared to other AI models, such as Bard, demonstrating the potential of AI models to match or even exceed human standards in tasks like processing and applying medical knowledge at a postgraduate level [7].

In the educational domain, ChatGPT has excelled in standardized tests such as the Test of Understanding in College Economics, scoring in the 91st to 99th percentile [8]. Furthermore, studies have highlighted ChatGPT's proficiency in various standardized admission tests in the UK, showcasing its potential as an innovative tool for education and test preparation [9]. The model has also shown capabilities in history exams and has been compared to students' scores, indicating a commendable level of proficiency in the subject [10].

As for the Arabic language, Alkaoud [11] introduces a new benchmark for evaluating large language models in English and Arabic. The author built an evaluation dataset based on the General Aptitude Test (GAT) to measure the linguistic capabilities of LLMs. The study demonstrates that ChatGPT-4's Arabic capabilities are significantly better than ChatGPT's.

In summary, while large language models show promising results on various exams, there remains a need for robust, natively developed Arabic benchmarks to rigorously evaluate mathematical reasoning and language understanding abilities tailored for the Arabic context.

## III. DATASET DESCRIPTION

The Qiyas exam includes two sections: quantitative and verbal, as mentioned in the previous section. All questions in both sections are of multiple choice with four choices for each question. The quantitative section comprises of questions to test students' intellectual abilities in math, geometry, algebra, and data analysis. The verbal section comprises of questions to test students' linguistic abilities in semantic relations, linguistic structures, and comprehension [3].

In the quantitative section, there are four types of questions, as outlined by [12]:
1) *Math:* Transforming verbal statements into solvable equations that involves basic arithmetic operations such as addition, subtraction, multiplication, and division.
2) *Geometry:* Applying geometric formulas and principles encompassing properties of triangles, area computations, angle measurements, and related concepts.
3) *Algebra:* Analyzing and resolving a set of algebraic equations or expressions to find the numerical value of an unknown variable, discern numerical sequences and patterns, among other related concepts.
4) *Statistics:* Applying fundamental principles in probability theory and statistics that involves utilizing mathematical concepts to analyze, interpret data, and make predictions.

In the verbal section, there are five types of questions:
1) *Reading Comprehension:* Comprehending reading passages and responding to questions that pertain to the content of the passage.
2) *Sentence Completion:* Extracting the appropriate word from the choices to complete a sentence with a missing word.
3) *Contextual Error:* Identifying the contextual discrepancy in the sentence and pinpointing the word whose meaning contradicts the overall meaning of the sentence (The error is not a spelling or grammar error).
4) *Verbal Analogy:* Recognizing the connection between the two words in the question, then evaluating them based on analogous choices provided.
5) *Anomalous Word:* Detecting the distinct word that is not related to the connected choices by a particular association.

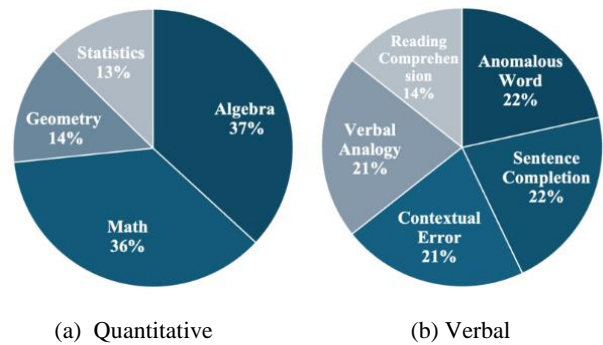

(a) Quantitative  (b) Verbal

Fig. 1. The distribution of Question types in the dataset in (a) for Quantitative and (b) for Verbal.

The appendix shows an example of each question type with its translation to English. The questions used in the evaluation were written by domain experts experienced in designing and grading Qiyas exams. Figure 1 shows the distribution of quantitative and verbal questions in the dataset, resulting in a total of **2,407** questions. In the quantitative section, the number of questions related to math and algebra surpasses those in geometry and statistics. The reason is that math and algebra questions do not necessitate reliance on charts or plots for answering. Unlike geometry and statistics questions, which often involve visual representations. We focused on questions that do not rely on visual representations, as indicated by a previous study [13], which revealed that ChatGPT-4 struggled to retain and process visual information, highlighting the necessity of adding image descriptions in the evaluation of ChatGPT-4. It is important to note that the Qiyas exam does not include image descriptions, emphasizing that the objective of the evaluation aims to mirror the examination process of students. On the other hand, the verbal section exhibits a relatively balanced distribution of question types, except for reading comprehension, which demonstrates a lower prevalence.

## IV. EVALUATION APPROACH

Our evaluation approach starts by formulating a prompt for each question within our dataset. The prompts used were the exact prompts in Arabic utilized in the official Qiyas exam as provided by authorized guides [12]. This approach also aligns with the examination methodology experienced by students, mitigating the risk of injecting our own subjective influences into the prompts. Due to ChatGPT's tendency to generate lengthy explanations for questions, which complicates the process of extracting the answer, we have introduced the instruction "Write the answer only" in the prompt. This measure is intended to ensure that only the answer is provided without additional explanation. While ChatGPT-4 complied with this directive, ChatGPT-3.5 persisted in including explanations in most answers, thus failing to adhere to the specified command.

In the evaluation phase, we employed OpenAI's API (`ChatGPT-3.5-turbo` and `ChatGPT-4`) to prompt and extract corresponding answers [14]. We initiated the evaluation of the models by employing zero-shot prompts, but we subsequently extended it by incorporating one-shot and 3-shot prompts. This adjustment was made to investigate the impact of varying prompt complexities on the model's performance and to explore how providing additional context influences the model's responses. The examples used in both the one-shot and 3-shot prompts remained consistent across all questions. Figure 2 provides an example of the prompt methodology used, noting that the prompts were originally in Arabic but translated to English for clarity purposes.

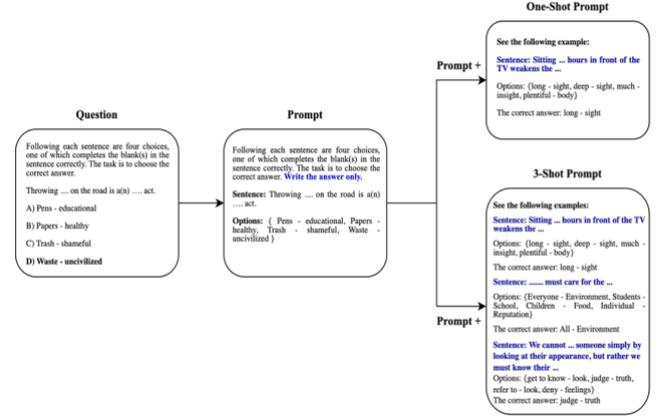

Fig. 2. Example of translated prompt questions with different prompt settings.

## V. RESULTS AND DISCUSSION

Table I displays the results of our experiments on ChatGPT-4 and ChatGPT-3.5-turbo. The evaluation metric used is the accuracy (See equation 1).

$$\text{Accuracy of Question Type A} = \frac{\text{Number of Correct Answers in A}}{\text{Total Number of Questions in A}} \times 100 \quad (1)$$

### A. Quantitative Section

In the quantitative section, ChatGPT-4 excelled in math and statistics with zero-shot prompts, indicating its strong capability in answering these types of questions without providing additional context. However, in geometry and algebra, ChatGPT-4 exhibited its peak performance with an accuracy of **81%** and **63%** respectively when presented with 3-shot prompts, suggesting that its capabilities were optimized when provided with more context, enabling it to leverage additional information to enhance its proficiency in these areas.

On the other hand, ChatGPT-3.5-turbo's accuracy did not surpass **65%** across all prompt settings. Its peak performance of **65%** accuracy was attained in geometry with the 3-shot prompt. Notably, ChatGPT-3.5-turbo achieved its best accuracy in statistics questions when provided with 3-shot prompts, contradicting ChatGPT-4's results for the same question type, where it excelled with zero-shot prompts. The contradictory results observed could be attributed to differences in their training data and model architectures. ChatGPT-4's larger training corpus and advanced architecture might enable it to leverage patterns and context more effectively in zero-shot settings, while ChatGPT-3.5-turbo benefits from additional contextual information provided in few-shot prompts.

### B. Verval Section

In the verbal section, ChatGPT-4 demonstrated notable proficiency in reading comprehension, achieving an accuracy peak of **80%** with one-shot and 3-shot prompts. This showcases that providing context and examples can positively influence the results in language-related tasks. However, it is worth noting that ChatGPT-4 also excelled in reading comprehension with zero-shot prompts, indicating its strong language understanding capabilities even without supplementary examples. We believe that these exceptional results were achieved due to the nature of the question and its dependency on the passage to draw the connections required to answer questions accurately. Following closely, sentence completion yielded an accuracy of **74%** with the 3-shot prompt.

Conversely, ChatGPT-3.5-turbo exhibited its highest accuracy across most question types when employing one-shot prompts, except for anomalous word, where it performed best with 3-shot prompt, and contextual error where it achieved the same accuracy for all prompt settings.

### C. Summary

The overall results show that ChatGPT-4 outperforms ChatGPT-3.5-turbo in a wide variety of linguistic and mathematical domains with a total average accuracy of **64%**, whereas GPT-3.5-turbo achieved an average total accuracy of **49%** in all prompt settings.

Compared to Alkaoud benchmark study on the Arabic language [11], our dataset size surpasses his study's dataset, which comprised of only **468** Arabic verbal questions, whereas our dataset comprised of **2,407** both quantitative and verbal questions. This larger dataset enables a more robust evaluation of these models' capabilities across different question types and prompts. Alkaoud followed a comparison approach between the Arabic and English language with zero-shot prompt setting, whereas we focused solely on the Arabic language with different prompt settings to evaluate the models' performance. In line with Alkaoud's findings, ChatGPT-4 demonstrated superior performance in reading comprehension, achieving **74%** accuracy, while our results achieved **77%** accuracy in the same question type with zero-shot prompts. In our experiments in the verbal section, the lowest accuracy achieved with zero-shot prompts was in the contextual error question type, reaching an accuracy of **56%**. In contrast, Alkaoud's research achieved a higher accuracy of **63.37%** in the same question type. We suspect that this variance in results could be attributed to Alkaoud utilizing a publicly accessible dataset, suggesting that ChatGPT might have been trained on it, while our dataset remains non-public.

TABLE I. EVALUATION RESULTS OF CHATGPT-4 AND CHATGPT-3.5-TURBO

| Section | Question Type | ChatGPT-4 | | | ChatGPT-3.5-turbo | | |
|---|---|---|---|---|---|---|---|
| | | *0-Shot* | *1-Shot* | *3-Shot* | *0-Shot* | *1-Shot* | *3-Shot* |
| **Quantitative** | Math | **65%** | 61% | 61% | **51%** | 45% | 48% |
| | Geometry | 61% | 80% | **81%** | 57% | 63% | **65%** |
| | Algebra | 54% | 61% | **63%** | 41% | **51%** | 48% |
| | Statistics | **71%** | 63% | 67% | 42% | 45% | **55%** |
| **Verbal** | Reading Comprehension | 77% | **80%** | 80% | 63% | **66%** | 64% |
| | Sentence Completion | 72% | 73% | **74%** | 44% | **73%** | 72% |
| | Contextual Error | 56% | **59%** | 43% | 41% | 41% | 41% |
| | Verbal Analogy | 58% | 58% | **59%** | 34% | **37%** | 35% |
| | Anomalous Word | **59%** | 59% | 58% | 47% | 46% | **49%** |
| | **Total** | 63% | **65%** | 63% | 46% | **51%** | 51% |
| | **Total Average** | | **64%** | | | **49%** | |

## VI. ERROR ANALYSIS

To gain a deeper understanding of the errors made by both models and to identify any patterns or common error types, we conducted a comprehensive error analysis on the zero-shot results for both ChatGPT-4 and ChatGPT-3.5-turbo. The analysis, generated with the assistance of ChatGPT-4o [14], aimed to categorize the errors and provide detailed insights into the specific challenges faced by each model. We also evaluated the performance of another language model, Gemini-pro by Google [15], on the questions that were incorrectly answered by ChatGPT-4 and ChatGPT-3.5-turbo to determine if alternative models could perform better. Table II showcases the zero-shots error analysis results with the most common error types for each section.

### A. Quantitative Section Error Analysis

In the quantitative section, Algebra questions resulted in the highest error rate for both models. Both models exhibited difficulty in solving complex equations for the missing variable. Additionally, they struggled with identifying the correct relationship (>, <, or =) between various algebraic expressions. This highlights the need for further development and incorporation of more diverse training data encompassing complex algebraic equations. Conversely, the models performed exceptionally well on problems involving simple and direct equations, suggesting that both models are adept at handling straightforward scenarios that lack complex transformations.

The analysis of statistical tasks revealed a more nuanced picture. ChatGPT-4 achieved a significantly lower error rate compared to ChatGPT-3.5-turbo. Notably, ChatGPT-3.5-turbo encountered specific difficulties with probability questions and problems involving combinatorics.

### B. Verbal Section Error Analysis

In the verbal section, Contextual Error questions resulted in the highest error rate for ChatGPT-4. Both models faced difficulty in differentiating between synonymous answer choices and comprehending the deeper context of the sentence. Our assumption suggests that synonyms might share similar statistical properties that may challenge the model in distinguishing the correct word in a specific context.

TABLE II. ERROR ANALYSIS RESULTS OF CHATGPT-4 AND CHATGPT-3.5-TURBO

| Section | Question Type | ChatGPT-4 | ChatGPT-3.5-turbo | Common Error Types |
|---|---|---|---|---|
| **Quantitative** | Math | 35% | 49% | Arithmetic Problems and Verbal Equations |
| | Geometry | 39% | 43% | Triangle and Angle Properties |
| | Algebra | **46%** | **59%** | Solving for the Missing Variable and Algebraic Comparison Questions |
| | Statistics | 29% | 58% | Multiple Conditions Probability Questions and Combinatorics problems |
| **Verbal** | Reading Comprehension | 23% | 37% | Inference and Reasoning Questions |
| | Sentence Completion | 28% | 56% | Misunderstanding of Sentence Structure and Context |
| | Contextual Error | 44% | 59% | Synonym Differentiation and Contextual Understanding |
| | Verbal Analogy | 42% | 66% | Selecting an Incorrect But Related Pair |
| | Anomalous Word | 41% | 53% | Cultural and Contextual Knowledge |

TABLE III. GEMINI-PRO EFFECTIVENESS ON INCORRECT ANSWERS FROM CHATGPT-4 & CHATGPT-3.5-TURBO

| Section | Question Type | # of Wrong Answers (ChatGPT-4) | Gemini-pro (ChatGPT-4) Accuracy | # of Wrong Answers (ChatGPT-3.5-turbo) | Gemini-pro (ChatGPT-3.5-turbo) Accuracy |
|---|---|---|---|---|---|
| **Quantitative** | Math | 128/370 | 30% | 180/370 | 36% |
| | Geometry | 55/142 | 44% | 61/142 | 36% |
| | Algebra | 168/374 | 27% | 221/374 | 29% |
| | Statistics | 37/128 | 19% | 74/128 | 43% |
| **Verbal** | Reading Comprehension | 46/199 | **46%** | 73/199 | **56%** |
| | Sentence Completion | 83/300 | 30% | 168/300 | 38% |
| | Contextual Error | 132/298 | 29% | 176/298 | 34% |
| | Verbal Analogy | 124/298 | 16% | 197/298 | 26% |
| | Anomalous Word | 116/300 | 30% | 160/300 | 39% |

On the other hand, Verbal Analogy questions resulted in the highest error rate for ChatGPT-3.5-turbo. These questions require identifying the closest relationship between two given words from a set of answer choices. The difficulty appears to stem from the inherent ambiguity within the answer choices themselves. Since each answer choice likely shares some form of connection to the original word pair, the model struggles to pinpoint the most precise analogy. Further research is needed to explore how LLMs can be better equipped to handle tasks that require reasoning about subtle semantic relationships between words.

### C. Gemini-pro Results

To evaluate the performance of other language models on the incorrectly answered questions by ChatGPT-4 and ChatGPT-3.5-turbo, we have provided the same questions to Gemeni-pro by Google [15] and its response were compared to the originally incorrect outputs from ChatGPT-4 and ChatGPT-3.5-turbo.

Table III summarizes the evaluation results. We can see that Gemini-pro demonstrated promising performance in correctly answering most questions. Notably, Gemini-pro excelled in the Reading Comprehension question type, suggesting a strong capability for leveraging relevant background information for response generation.

However, Gemini-pro's performance on verbal analogy questions was lower. Verbal analogy questions demand the model to grasp the relationship between word pairs and identify another pair with a similar connection. This task can be challenging for LLMs, as it necessitates not only understanding individual word meanings but also the intricate ways words can relate to each other. Interestingly, both ChatGPT-4 and ChatGPT-3.5-turbo also exhibited lower performance on this question type, potentially indicating a general limitation in current LLM technology.

It is worth noting that Gemini-pro showed strength in following instructions by responding with the answer only without explanation or additional context. Unlike ChatGPT-4 and ChatGPT-3.5-turbo, which occasionally included extraneous information in their responses, Gemini-pro consistently provided only the answer to the question, as instructed.

The result of this evaluation suggests that Gemini-pro is a promising LLM that shows particular strength in the Arabic language by being able to correctly answer a wide range of questions that were incorrectly answered by ChatGPT-4 and ChatGPT-3.5-turbo.

### VII. CONCLUSION AND FUTURE WORK

This research paper introduces the Qiyas benchmark, a novel evaluation framework developed to comprehensively assess the mathematical reasoning and language understanding capabilities of large language models (LLMs) in the Arabic language. The Qiyas benchmark is a standardized General Aptitude Test (GAT) used for university admissions in Saudi Arabia, ensuring its quality and relevance to real-world assessment.

The key findings of this paper are:
1) ChatGPT-4 outperformed ChatGPT-3.5-turbo across both the quantitative (math) and verbal (language) sections of the benchmark. This suggests that the newer, more advanced model has made notable progress in Arabic language understanding and mathematical reasoning compared to its predecessor.
2) The performance of the models varied depending on the prompt setting (zero-shot, one-shot, 3-shot). In general, providing more contextual information through one-shot and 3-shot prompts improved the models' accuracy, particularly in the verbal section tasks like reading comprehension.
3) The results highlight the current limitations of state-of-the-art LLMs in handling the complexities of the Arabic language, including its unique morphology and writing system. This underlines the need for more Arabic-focused training data and model development efforts to enhance the mathematical and linguistic capabilities of future Arabic LLMs.

The release of the Qiyas benchmark represents a significant contribution to the field, as it provides a robust, standardized evaluation framework for assessing the capabilities of Arabic language models. This resource can drive the development of more capable Arabic LLMs by serving as a benchmark for progress and identifying specific areas requiring further research and improvement. Future work includes expanding the dataset to include image-based questions, enabling the evaluation of multimodal models' ability to integrate language and visual understanding for Arabic-based tasks. Additionally, assessing a wider range of state-of-the-art LLMs on the Qiyas benchmark will provide a more comprehensive understanding of the current capabilities and limitations of Arabic language AI systems. Overall, this work lays the foundation for advancing the state-of-the-art in Arabic language understanding and reasoning for large language models.

TABLE IV. EXAMPLE OF EACH QUESTION IN THE DATASET

| Section | Question Type | Question | Translated Question |
|---|---|---|---|
| Quantitative | Math | إذا كان ثمن 6 دفاتر يعادل 12 ريال فكم يكون ثمن 9 دفاتر من نفس النوع؟<br>أ) ١٢   **ب) ١٨**<br>ج) ٢٤   د) ٢٧ | If the price of 6 notebooks is equivalent to 12 riyals, how much is the price of 9 notebooks of the same type?<br>A) 12   **B) 18**<br>C) 24   D) 27 |
| | Geometry | مستطيل محيطه 40 سم وطوله يزيد عن عرضه بمقدار 2 فما هي مساحته؟<br>**أ) ٩٩ سم**   ب) ٤٠٠ سم<br>ج) ٢٠ سم   د) ١٠٨ سم | A rectangle has a perimeter of 40 cm, and its length is 2 times its width. What is its area?<br>**A) 99 cm**   B) 400 cm<br>C) 20 cm   D) 108 cm |
| | Algebra | إذا كان س$^2$ - ص$^2$ = 16 وكان س - ص = 2 فإن س + ص = .......<br>**أ) ٨**   ب) ٢<br>ج) ٤   د) ١ | If $x^2 - y^2 = 16$ and $x - y = 2$, then $x + y = $ .......<br>**A) 8**   B) 2<br>C) 4   D) 1 |
| | Statistics | عند رمي مكعب أرقام مرقم من 1 إلى 6 فما هو احتمال أن يكون الوجه العلوي عدد أولي؟<br>**أ) ٢/١**   ب) ٣/١<br>ج) ٦/١   د) ٤/١ | When throwing a cube numbered from 1 to 6, what is the probability that the top side is a prime number?<br>**A) 1/2**   B) 1/3<br>C) 1/6   D) 1/4 |
| Verbal | Reading Comprehension | الأسئلة التالية تتعلق بالنص الذي يسبقها، بعد كل سؤال أربع اختيارات، أحدها صحيح. المطلوب هو قراءة النص بعناية، واختيار الإجابة الصحيحة عن كل سؤال.<br>الحمية القاسية<br>هناك من يمارس الحمية القاسية دون أن يكتسب العادات الغذائية الجيدة لذا يعود للسمنة مجدداً<br>يتحدث النص عن:<br>أ) الحمية القاسية لا تؤدي لتخفيف الوزن<br>ب) الحمية القاسية نادراً ما تنتهي بالفشل<br>ج) الحمية تفيد الجسم حتى مع العادات الغذائية السيئة<br>**د) غياب العادات الغذائية الجيدة يقلل من فائدة الحمية** | The following questions relate to the preceding text. After each question, there are four choices, one of which is correct. The task is to read the text carefully and select the correct answer for each question.<br>The Harsh Diet<br>Some people follow a harsh diet without acquiring good eating habits, so they return to obesity again.<br>The text is talking about:<br>A) Harsh diets do not lead to weight loss.<br>B) Harsh diets rarely end in failure.<br>C) Diet benefits the body even with bad eating habits.<br>**D) Lack of good eating habits reduces the effectiveness of the diet** |
| | Sentence Completion | تلي كل جملة من الجمل الآتية أربع اختيارات، أحدهما يكمل الفراغ أو الفراغات في الجملة إكمالاً صحيحاً. المطلوب هو اختيار الإجابة الصحيحة.<br>رمي ....... في الطريق فعل غير ......<br>أ) الأقلام - تربوي   ب) الأوراق - صحي<br>ج) القاذورات - مُشين   **د) المخلفات – حضاري** | Following each sentence are four choices, one of which completes the blank(s) in the sentence correctly. The task is to choose the correct answer.<br>Throwing .... on the road is a(n) .... act.<br>A) Pens - educational   B) Papers - healthy<br>C) Trash - shameful   **D) Waste - uncivilized** |
| | Contextual Error | لكل جملة مما يأتي أربع خيارات. المطلوب هو: تحديد الكلمة التي لا يتفق معناها مع المعنى العام للجملة. (الخطأ ليس إملائياً ولا نحوياً)<br>شخصية الفرد لها عدة صفات جسمية مثل حسن الهيئة وعقلية مثل التفكير ونفسية مثل الصدق والطول<br>أ) التفكير   **ب) الطول**<br>ج) جسمية   د) شخصية | For each of the following sentences, there are four choices. The task is to identify the word that does not match its meaning with the overall meaning of the sentence. (The error is not related to spelling or grammar).<br>The individual's character has several physical traits like good physique, mental traits like thinking, and psychological traits like honesty and height.<br>A) Thinking   **B) Height**<br>C) Physical   D) Character |
| | Verbal Analogy | في بداية كل سؤال مما يأتي، كلمتان ترتبطان بعلاقة معينة، تتبعهما أربعة أزواج من الكلمات، أحدها ترتبط فيه الكلمتان بعلاقة مشابهة للعلاقة بين الكلمتين في بداية السؤال. المطلوب، هو: اختيار الإجابة الصحيحة<br>بناء:هدم<br>**أ) قبل:بعد**   ب) ندم:حزن<br>ج) نعاس:نوم   د) هدوء:سكون | At the beginning of each question from the following, there are two words related in a certain way, followed by four pairs of words, one of which has a relationship similar to the relationship between the two words at the beginning of the question. The task is to choose the correct answer.<br>Construction: Destruction<br>**A) Before: After**   B) Regret: Sadness<br>C) Drowsiness: Sleep   D) Quietness: Stillness |
| | Anomalous Word | الاسئلة الآتية يتضمن كل منها أربع كلمات، يجمع ثلاثاً منها رابط معين وواحدة مختلفة عنها. المطلوب هو تحديد الكلمة المختلفة.<br>أ) حب   **ب) رضا**<br>ج) عشق   د) هيام | The following questions each contain four words, three of which share a certain link and one is different. The task is to identify the different word.<br>A) Love   **B) Satisfaction**<br>C) Passion   D) Infatuation |